\documentclass[conference]{IEEEtran}
\IEEEoverridecommandlockouts
\usepackage{cite}
\pdfoutput=1
\usepackage{amsmath,amssymb,amsfonts}
\usepackage{algorithm, algorithmic}
\usepackage{graphicx}
\usepackage{textcomp}
\usepackage{xcolor}
\def\BibTeX{{\rm B\kern-.05em{\sc i\kern-.025em b}\kern-.08em
		T\kern-.1667em\lower.7ex\hbox{E}\kern-.125emX}}
\usepackage{enumerate}
\usepackage{pageno}
\usepackage{url}
\usepackage[utf8]{inputenc}
\DeclareUnicodeCharacter{2212}{-}
\begin{document}
	
	\title{A Multi-operator Ensemble LSHADE with Restart and Local Search Mechanisms for Single-objective Optimization}
	\author{\IEEEauthorblockN{Dikshit Chauhan$^{1*}$, Anupam Trivedi$^{2}$, Shivani$^{1}$}
		
		\thanks{$^{1*}$ Department of Mathematics and Computing, Dr. B.R. Ambedkar National Institute of Technology Jalandhar, Jalandhar-144008, Punjab, India, \textit{e-mail:} dikshitchauhan608@gmail.com and sainishivani2310@gmail.com}
		\thanks{$^{2}$ Department of Electrical and Computer Engineering of the National University of Singapore, Singapore, \textit{e-mail:} eleatr@nus.edu.sg}
	}
	\maketitle
	
	\begin{abstract}
In recent years, multi-operator and multi-method algorithms have succeeded, encouraging their combination within single frameworks. Despite promising results, there remains room for improvement as only some evolutionary algorithms (EAs) consistently excel across all optimization problems. This paper proposes mLSHADE-RL, an enhanced version of LSHADE-cnEpSin, which is one of the winners of the CEC 2017 competition in real-parameter single-objective optimization. mLSHADE-RL integrates multiple EAs and search operators to improve performance further. Three mutation strategies such as \textit{DE/current-to-pbest-weight/1 with archive}, \textit{DE/current-to-pbest/1 without archive}, and \textit{DE/current-to-ordpbest-weight/1} are integrated in the original LSHADE-cnEpSin. A restart mechanism is also proposed to overcome the local optima tendency. Additionally, a local search method is applied in the later phase of the evolutionary procedure to enhance the exploitation capability of mLSHADE-RL. mLSHADE-RL is tested on 30 dimensions in the CEC 2024 competition on single objective bound constrained optimization, demonstrating superior performance over other state-of-the-art algorithms in producing high-quality solutions across various optimization scenarios.
	\end{abstract}
	
	
\section{Introduction}
Evolutionary algorithms (EAs), a subset of evolutionary computation, solve one or more objective functions by searching for the global optimum solution in $\mathbb{R}^n$. This can be formalized as:\begin{equation}
	\text{Find}~x_{min}\in \mathcal{S}~\forall~x\in\mathcal{S},f_v(x_{min})\leq f_v(x),
\end{equation} where $\mathcal{S}\in\mathbb{R}$ is a bounded set and $f_v:\mathcal{S}\rightarrow\mathbb{R}$ is an $n-$dimensional real-valued function. The function $f_v$ must be bounded but need not be differentiable or continuous. EAs are effective and efficient for global optimization, making them suitable for many applications, such as attacking neural networks, forecasting energy consumption, and predicting financial models. They are known for their flexibility and robustness in handling complex optimization problems.

Among EAs, differential evolution (DE)~\cite{storn1997differential} has gained popularity for solving continuous optimization problems and has demonstrated superiority over other methods in various complex scenarios. Numerous mutation strategies have been developed to enhance DE's performance. However, no single strategy consistently outperforms others across all problem types, highlighting the need for diverse approaches within DE frameworks.

Many algorithms have incorporated multiple strategies, operators, or parameter adaptation techniques to address these limitations into a single framework. These multi-phase, multi-operator, and ensemble-based approaches employ a selection mechanism to dynamically choose the best-performing strategy or operator during the search process~\cite{das2016recent}. Despite their advantages, these approaches still present research opportunities for further improvement. In multi-operator-based algorithms, several notable contributions include Qin et al.~\cite{qin2008differential} introduced the self-adaptive DE algorithm (SaDE), which utilizes four mutation strategies. Each individual in the population is assigned to one strategy based on a given probability, which is updated according to the strategy's success and failure rates during previous generations.

Two mutation strategies are introduced in jDE~\cite{zamuda2012population}, with the population size adaptively reduced during the evolutionary process. It was later extended to include a self-adaptation mechanism for parameter control~\cite{brest2013real}, utilizing three DE strategies, each applied to specific groups of individuals based on predefined parameters. Individuals who had not improved over several generations were reinitialized with a predefined probability. These algorithms utilize multiple mutation strategies sequentially, applying each strategy for a predefined number of generations. Additionally, they incorporate mechanisms to reduce the population size dynamically. These advancements illustrate the evolution of DE algorithms, combining multiple strategies to enhance performance across diverse optimization problems~\cite{pant2020differential}. However, ongoing research continues to refine these approaches to achieve more robust and universally effective optimization solutions. One approach to designing an improved algorithm is to combine multiple algorithms and operators within a single framework, discussed in Section~\ref{subsec: multi operator DE}. These unified methods have shown significant improvements in solving various optimization problems.

Inspired by multiple EAs, we propose a \textbf{M}odified version of \textbf{LSHADE}-cnEpSin~\cite{awad2016ensemble} with \textbf{R}estart mechanism and \textbf{L}ocal search, called mLSHADE-RL, which incorporates a multi-operator approach. The population is updated using three different mutation strategies. To address the tendency of DE to get stuck in local optima, a restart mechanism with horizontal and vertical crossovers is used to track stagnation and replace less promising vectors. Additionally, a local search method is applied in the later phase of the evolutionary procedure to enhance the exploitation capability of mLSHADE-RL.

The remainder of this paper is organized as follows: Section~\ref{sec: related works} discusses the original DE, its components, and associated works. The design elements of the proposed mLSHADE-RL are explained in Section~\ref{sec: proposed algorithm}. Section~\ref{sec: experimental} presents the experimental results. Concluding remarks are provided in Section~\ref{sec: conclusion}.
	
	\section{Related Works}\label{sec: related works}
	This section explains the original DE and its variants used to design the proposed algorithm.
	\subsection{Differential Evolution}\label{subsec: de}
	This section briefly explains differential evolution (DE)~\cite{storn1997differential}, covering all its design elements. The population in DE, also known as DE/rand/1/bin, is denoted as $PS$ and consists of $N_p$ individuals. Each individual, referred to as a target or parent vector, is represented by a solution vector $x_i=\{x_{i,1},x_{i,2},\ldots,x_{i,D}\}$ where $D$ is the dimension of the search space. Throughout the evolutionary process, individuals are updated over generations $G=1,2,\ldots,G_{max}$, where $G_{max}$ is the maximum number of generations. The $i^{th}$ individual in $PS$ is initialized randomly within the search bounds ($x_{lb}=\{x_{1,lb},x_{2,lb},\ldots,x_{N_p,lb}\}$ and $x_{ub}=\{x_{1,ub},x_{2,ub},\ldots,x_{N_p,ub}\}$) at generation $G$. This initialization is represented as $x_i^G=\{x_{i,1}^G,x_{i,2}^G,\ldots,x_{i,D}^G\}$. 
	
	Following initialization, DE employs operators such as mutation and crossover to evolve the individuals in $PS$, generating trial vectors. A comparison between these trial vectors and the target vectors, known as selection, determines which individuals will survive to the next generation. The detailed steps of DE are discussed as follows:\begin{enumerate}[(i)]
		\item \textbf{Mutation:} For each target vector $x_i^G$ at generation $G$, DE creates a mutant vector $v_{i}^{G+1}=\{v_{i,1}^{G+1},v_{i,2}^{G+1},\ldots,v_{i,D}^{G+1}\}$ using the following mutation process:\begin{equation}
		DE/rand/1:~~	v_{i}^{G+1}=x_{r_1}^G+F\cdot(x_{r_2}^G-x_{r_3}^G),
		\end{equation}   
		where $r_j$ (for $j=1,2,3$) are the mutually exclusive, randomly selected indices from the range $[1,N_p]$, and different to the base index $i$. The parameter $F$ is a scaling factor, which controls the influence of the difference vector $(x_{r_2}^G-x_{r_3}^G)$ and ranges between 0 and 2. A higher value of $F$ promotes exploration, while a lower value promotes exploitation.
		
		\item \textbf{Crossover:} After generating the mutant individual $v_i^{G+1}$ through mutation, the crossover operation is applied. In crossover, the mutant vector $v_i^{G+1}$ exchanges components with the target individual $x_i^{G}$ with a probability $CR \in[0,1]$ to form the trial individual $u_i^{G+1} = \{u_{i,1}^{G+1},u_{i,2}^{G+1},\ldots,u_{i,D}^{G+1}\}$. Two types of crossover are used in DE: $binomial$ and $exponential$. Here, we elaborate on the binomial crossover, which is preferred over exponential. In binomial crossover, the target vector is combined with the mutant vector according to the following condition:
		\begin{equation}\label{eq: binomial crossover}
		\small
		u_i^{G+1} = \begin{cases}
		v_i^{G+1},&{\text{if}~\left( {rand_{i,j}(0,1) \le CR{\rm{\;or\;}}j = {j_{rand}}} \right)}\\
		x_i^{G},&\text{otherwise},
		\end{cases}
		\end{equation}
		where $rand_{i,j}(0,1)$ is a uniformly distributed random number and ${j_{rand}} \in \{1,2,. . .,D\}$ is a randomly chosen index that ensures the trial individual inherits at least one component from the mutant individual. In this equation, the crossover rate $CR$ controls the number of components inherited from the mutant vector.
		
		\item \textbf{Selection:} In the selection procedure, DE employs a greedy selection scheme to determine which vector will survive. In each generation, if the trial vector $u_i^{G+1}$ is fitter than the target vector $x_i^{G}$, the trial vector will replace the target vector. Otherwise, the target vector $x_i^{G}$ is retained. For the minimization case, the selection strategy is defined as follows:\begin{equation}\label{eq: de selection}
		x_i^{G+1}=\begin{cases}
		u_i^{G+1},&\text{if}~f_v(u_i^{G+1})<f_v(x_i^{G}),\\
		x_i^{G},&\text{otherwise}.
		\end{cases}
		\end{equation}
	\end{enumerate}
	
	\subsection{Unified DE}
	The unified DE (UDE), proposed by Trivedi et al.~\cite{trivedi2017unified}, was one of the winners in the CEC 2017 constrained optimization competition. Inspired by JADE, CoDE, SaDE, and the ranking-based mutation operator, UDE employs three trial vector generation strategies similar to CoDE: \textit{rank-DE/rand/1/bin}, \textit{rank-DE/current-to-rand/1}, and \textit{rank-DE/current-to-pbest/1}. UDE used the dual subpopulations framework: 
	the top subpopulation and the bottom subpopulation, each with distinct learning mechanisms. The top subpopulation employs all three trial vector generation strategies to update each generation's target vector $x_{i}^G$. In contrast, the bottom subpopulation used strategy adaptation, periodically self-adapting its trial vector $u_i^G$ generation strategies based on experiences and insights from generating promising solutions in the top subpopulation. Additionally, UDE incorporates a local search operation-based mutation to enhance its performance further. 
	\subsection{SHADE}
	The success history-based adaptive DE (SHADE)\cite{tanabe2013success} extends JADE\cite{zhang2009jade}. In JADE, each individual updates using the \textit{DE/current-to-pbest/1} mutation strategy:
	\begin{equation}
	v_i^G=x_i^G+F_i^G\cdot (x_{pbest}^G-x_i^G)+F_i^G\cdot (x_{r_1}^G-x_{r_2}^G),
	\end{equation} where $pbest$ is selected from the top $100p\%$ of the population. Each $x_i^G$ has its control parameters $F_i$ and $CR_i$, generated probabilistically from adaptive parameters $\mu_F$ and $\mu_{CR}$ using Cauchy and normal distributions. Successful $CR_i$ and $F_i$ values that improve trial vectors are recorded in $S_{CR}$ and $S_F$. At the end of each generation, the arithmetic mean of $S_{CR}$ and the Lehmer mean of $S_F$ update $\mu_{CR}$ and $\mu_F$.

	SHADE enhances this by leveraging a historical record of successful parameter settings. It maintains memory for $F$ and $CR$ parameters ($M_F$ and $M_{CR}$) with size $H$. In each generation, control parameters $CR_i$ and $F_i$ for each individual $x_i^G$ are generated from $M_{CR,r_i}$ and $M_{F,r_i}$, using normal and Cauchy distributions, where $r_i$ is a randomly selected index from $[1, H]$. Successful $CR_i$ and $F_i$ values that generate better trial vectors are recorded in $S_{CR}$ and $S_F$. At the end of each generation, the weighted arithmetic mean of $S_{CR}$ and the weighted Lehmer mean of $S_F$ update $M_{CR,j}$ and $M_{F,j}$, respectively, where $j$ determines the position to be updated in the memory.
	\subsection{EBSHADE}\label{subsec: ebshade}
	In EBSHADE, a more greedy mutation strategy, \textit{DE/current-to-ordpbest/1}, is proposed to enhance exploitation capability~\cite{mohamed2019novel}. This strategy orders three selected vectors from the current generation to perturb the target vector, using directed differences to mimic gradient descent and direct the search toward better solutions. In \textit{ordpbest}, one vector is selected from the global top $p$ best vectors, and the other two are randomly chosen from the entire population. These three vectors are then ordered by their fitness values: the best is $x_{ordpbest}^G$, the median is $x_{ordpmed}^G$, and the worst is $x_{ordwst}^G$. The trail vector $v_i^G$ is generated by the target vector $x_i^G$ and those above three ordered vectors, as follows:
	\begin{equation}
	v_i^G=x_i^G+F\cdot (x_{ordpbest}^G-x_i^G)+F\cdot (x_{ordpmed}^G-x_{ordpwst}^G)
	\end{equation}
	\subsection{L-SHADE}\label{subsec: lshade}
	L-SHADE~\cite{tanabe2014improving}, a winner of CEC 2014 single-objective optimization problems, extends SHADE with linear population size reduction (LPSR), continuously decreasing the population size according to a linear function. Initially, the population size is $N_p^{init}$, and it reduces to $N_p^{min}$ by the end of the run. After each generation $G$, the next generation's population size $N_p^{G+1}$ is calculated as:\begin{equation}\label{eq: linear reduction pop}
	N_p^{G+1}=round\left[N_p^{init}+\left(\frac{N_p^{min}-N_p^{init}}{nfe_{max}}\right)\cdot nfe\right],
	\end{equation} where $nfe$ and $nfe_{max}$ are the current and maximum number of evaluations. If $N_p^{G+1}$ is smaller than $N_p^{G}$, the $(N^G - N^{G+1})$ individuals in population are removed from the bottom.
	\subsection{LSHADE-EpSin}\label{subsec: lshade-epsin}
	To enhance L-SHADE's performance, LSHADE-EpSin introduces an ensemble approach to adapt the scaling factor using an efficient sinusoidal scheme~\cite{awad2016ensemble}. This method combines two sinusoidal formulas: a non-adaptive sinusoidal decreasing adjustment and an adaptive history-based sinusoidal increasing adjustment. Additionally, a local search method based on Gaussian Walks is used in later generations to improve exploitation. In LSHADE-EpSin, a new mutation strategy, \textit{DE/current-to-pbest/1} with an external archive, is defined as:\begin{equation}
	v_i^G=x_i^G+F_i^G\cdot (x_{pbest}^G-x_i^G)+F_i^G\cdot (x_{r_1}^G-x_{r_2}^G),
	\end{equation}where $r_1\neq r_2\neq i$, $x_{r_1}^G$ is randomly selected from the whole population, and $x_{r_2}^G$ is randomly chosen from the union of the whole population $PS$ and an external archive $A$ ($A$ stores the inferior solutions recently replaced by offspring) (i.e., $x_{r_2}^G\in PS^G\cup A^G$).
	
	\textit{Parameter adaptation:} An ensemble of parameter adaptation methods is proposed to adjust the scaling factor $F$. The two strategies are: \begin{equation}\label{eq: ensemble lshade-epsin}
	\begin{cases}
	F_i^G=0.5\cdot\left(\sin(\pi(2\cdot f_q\cdot G+1))\cdot \frac{G_{max}-G}{G_{max}}+1\right),\\
	F_i^G=0.5\cdot\left(\sin(\pi(2\cdot f_{q_1}\cdot G+1))\cdot \frac{G}{G_{max}}+1\right),
	\end{cases}
	\end{equation} where $f_q$ and $f_{q_1}$ are frequencies of the sinusoidal function, with $f_q$ is fixed and $f_{q_1}$ adapted each generation using a Cauchy distribution:\begin{equation}
	f_{q_1}=randc(\mu f_{r_i},0.1),
	\end{equation}
	where $\mu f_{r_i}$ is Lehmer mean, randomly chosen from the external memory $M_{f_q}$ that stores the successful mean frequencies from previous generations in $S_{f_q}$. The index $r_i$ is selected from $[1,H]$ at the end of each generation. Both sinusoidal strategies are used in the first half of the generations. In the latter half, the scaling factor $F_{i}^G$ is updated using Cauchy distribution:\begin{equation}
	F_i^G=randc(\mu F_{r_i},0.1).
	\end{equation} Additionally, the crossover rate $CR_i^G$ is adapted throughout the evolutionary process using a normal distribution: \begin{equation}
	CR_i^G=randn(\mu CR_{r_i},0.1).
	\end{equation}

	\subsection{LSHADE-cnEpSin}\label{subsec: lshade_cnepsin}
	LSHADE-EpSin was introduced to enhance L-SHADE's performance using an adaptive ensemble of sinusoidal formulas for scaling factor $F$ effectively. It randomly selects one of two sinusoidal formulas in the first half of the generations: a non-adaptive sinusoidal decreasing adjustment or an adaptive sinusoidal increasing adjustment. This process is enhanced with a performance adaptation scheme to effectively choose between the two formulas. LSHADE-cnEpSin~\cite{awad2017ensemble} is the improved version of LSHADE-EpSin, which is used as a practical selection for scaling parameter $F$ and a covariance matrix learning with Euclidean neighborhoods to optimize the crossover operator. 
	
	\textit{Effective selection:} One of the two sinusoidal strategies is chosen based on their performance in previous generations. During a learning period $L_p$, the number of successful and discarded trial vectors generated by each configuration is recorded as $nS_j^G$ and $nF_j^G$, respectively. For the first $L_p$ generations, both strategies have equal probability $P_j$ and are chosen randomly. Afterward, probabilities are updated as follows:\begin{align}\label{eq: effective selection}
	\begin{cases}
	S_j^G=\frac{\sum_{i=G-L_p}^{G-1}nS_{i,j}}{\sum_{i=G-L_p}^{G-1}nS_{i,j}+\sum_{i=G-L_p}^{G-1}nF_{i,j}}+\epsilon,\\
	P_j=\frac{S_j^G}{\sum_{j=1}^JS_j^G},
	\end{cases}
	\end{align}where $S_j^G$ represents the success rate of the trial vectors generated by each sinusoidal strategy.
	
	\textit{New crossover operator:} In LSHADE-cnEpSin, the crossover operator uses covariance matrix learning with a Euclidean neighborhood (CMLwithEN) with probability $P_c$. Individuals are sorted by fitness, and the Euclidean distance from the best individual, $x_{best}$, is computed. A neighborhood region is formed around $x_{best}$ using the top $N_p\times P_s$ individuals (here $P_s$ is 0.5). As the population size decreases, so does this neighborhood size. The covariance matrix $C$ is computed from this region:\begin{equation}
	C=O_bD_g O_b^T,
	\end{equation}where $O_b$ and $O_b^T$ are the orthogonal matrices, and $D_g$ is the diagonal matrix with eigenvalues. The target and trial vectors are updated using the orthogonal matrix $O_b^T$:
	\begin{equation}
	x_i^{',G}=O_b^Tx_i^G,~~v_i^{',G}=O_b^Tv_i^G.
	\end{equation} The binomial crossover (Eq.~\eqref{eq: binomial crossover}) is applied to the updated vectors to create the trial vector $u_i^{',G}$. This trial vector is then transformed back to the original coordinate system:\begin{equation}
	u_i^G=O_bu_i^{',G}.
	\end{equation} 
	\subsection{Multi-operator DE}\label{subsec: multi operator DE}
	The performance of DE operators can vary throughout the evolutionary process and across different problems. This variability has led to developing multi-operator DE approaches, which adaptively emphasize better-performing operators at various stages. Single-operator DE variants often struggle to solve optimization problems effectively, prompting significant interest in multi-operator DE variants~\cite{elsayed2011multi}.
	
	Sallam et al.~\cite{sallam2017landscape} proposed an adaptive operator selection (AOS) method that uses the performance history of DE operators and the function's landscape to automatically select the most suitable DE strategy from a pool. This approach has shown superiority on 45 unconstrained optimization problems from the CEC2014 and CEC2015 competitions. Elsayed et al.~\cite{elsayed2011multi} introduced the self-adaptive multi-operator DE (SAMO-DE) algorithm, which dynamically emphasizes the best-performing DE variants based on fitness quality and constraint violations, outperforming state-of-the-art algorithms.
	
	Tasgetiren et al.~\cite{tasgetiren2010ensemble} developed an ensemble of discrete DE algorithms to improve performance by selecting appropriate parameter sets. Wu et al. \cite{wu2018ensemble} proposed a multi-population framework to derive potential DE variants based on their ensemble characteristics, using a reward sub-population scheme to prioritize the best-performing variants, demonstrating effectiveness on standard optimization problems.
	
	Chen et al. \cite{chen2019united} created a multi-operator DE algorithm incorporating an interior-point method for efficient evolutionary searches. Elsayed et al. \cite{elsayed2014testing} introduced the united multi-operator EA (UMOEA), which divides the population into multiple subpopulations, each evolving with different UMOEAs. The best-performing subpopulations are selected adaptively, while underperforming ones are updated via information exchange, leading to effective searches. This approach led to the development of an improved version, UMOEAsII \cite{elsayed2016testing}, achieving competitive results. Inspired by UMOEAs-II, an effective butterfly algorithm with CMA-ES was proposed and emerged as the winner of the CEC 2017 competition~\cite{kumar2017improving}.

	\begin{algorithm}
		\caption{Working procedure of mLSHADE-RL}\label{algo: proposed algorithm}
		\begin{algorithmic}[1]
			\STATE Initialize the population size $N_p$, $G=0$, $nfes_{max}=10000\cdot D$, $P_{LS}=0.1$, $P_{MS_1}=P_{MS_2}=P_{MS_3}=1/3$, $count=0$, and all other required parameters,
			\STATE Generate the initial population $PS$ within the search boundaries using Eq.~\eqref{eq: initial population} (Section~\ref{subsec: population initialization}),
			
			\WHILE{$nfes\leq nfes_{max}$}
			\STATE $G=G+1$,	

			\STATE Apply mLSHADE-cnEpSin to update $PS$ using Algorithm~\ref{algo: framework of mLSHADE-cnEpSin} (Section~\ref{subsec: mlshade-cnepsin}), and sort $PS$,
			\STATE Record counters and calculate the diversity metrics,
	 		\STATE If individuals are stagnated, use restart mechanism using Algorithm~\ref{algo: restart mechanism} (Section~\ref{subsec: restart mechanism}),
			\IF{$nfes\geq 0.85\cdot nfes_{max}$}
			\STATE Apply local search operator using Algorithm~\ref{algo: local search} (Section~\ref{subsec: local search}),
			\ENDIF
			\STATE $nfes=nfes+N_{p}.$
			\ENDWHILE
		\end{algorithmic}
	\end{algorithm}
	\section{Proposed Algorithm (mLSHADE-RL)}\label{sec: proposed algorithm}
	The working procedure of the proposed algorithm is explained in Algorithm~\ref{algo: proposed algorithm}. A random population of size $N_p$ is generated, with each individual within the search boundaries (Section~\ref{subsec: population initialization}). The population $PS$ is evolved using a \textit{LSHADE-cnEpSin with multi-operator} (mLSHADE) framework (Algorithm~\ref{algo: framework of mLSHADE-cnEpSin} of Section~\ref{subsec: mlshade-cnepsin}), and a restart mechanism with horizontal and vertical crossovers is incorporated in mLSHADE framework (Section~\ref{subsec: restart mechanism}). To enhance exploitation capability, a local search method, the sequential quadratic programming (SQP) method, is applied in the later stages with a dynamic probability (Section~\ref{subsec: local search}). This process repeats until a stopping criterion is met. The detailed algorithm components are discussed in the following subsections.
	\subsection{Population Initialization}\label{subsec: population initialization}
	The $i^{th}$ individual in $PS$ is initialized randomly within the search bounds ($x_{lb}=\{x_{1,lb},x_{2,lb},\ldots,x_{D,lb}\}$ and $x_{ub}=\{x_{1,ub},x_{2,ub},\ldots,x_{D,ub}\}$) at generation $G$. This initialization is represented as $x_i^G=\{x_{1,i}^G,x_{2,i}^G,\ldots,x_{D,i}^G\}$ and defined as:\begin{align}\label{eq: initial population}
	x_{i,j}^G=&x_{i,lb}+(x_{i,ub}-x_{i,lb})\cdot rand(1,N_p),\\\nonumber&
	i=1,2,\ldots,N_p, \text{and}~j=1,2,\ldots,D,
	\end{align}where $rand$ is a uniformly distributed real number between 0 and 1. An external archive $A$ is initialized by the initial population $x_i^G$.
	
	\subsection{LSHADE-cnEpSin with Multi-operator (mLSHADE)}\label{subsec: mlshade-cnepsin}
	In mLSHADE, three mutation strategies such as \textit{DE/current-to-pbest-weight/1 with archive}, \textit{DE/current-to-pbest/1 without archive}, and \textit{DE/current-to-ordpbest-weight/1} are used in the evolutionary procedure. The following DE mutation strategies ($MS_1$ to $MS_3$) to evolve the population $PS^G$ of size $N_{p}$.\begin{enumerate}[(i)]
		\item \textit{DE/current-to-pbest-weight/1 with archive} $(MS_1)$:\begin{equation}
		v_i^G=x_i^G+F_{i,w}^G\cdot (x_{pbest}^G-x_i^G)+F_{i}^G\cdot(x_{r_1}^G-x_{r_2}^G),
		\end{equation}
		\item \textit{DE/current-to-pbest/1 without archive} $(MS_2)$:\begin{equation}
		v_i^G=x_i^G+F_{i}^G\cdot (x_{pbest}^G-x_i^G+x_{r_1}^G-x_{r_3}^G),
		\end{equation}
		\item \textit{DE/current-to-ordpbest-weight/1} $(MS_3)$\begin{equation}
		v_i^G=x_i^G+F_{i,w}^G\cdot (x_{ordpbest}^G-x_i^G+x_{ordm}^G-x_{ordw}^G),
		\end{equation}
	\end{enumerate}
	here $r_1\neq r_2\neq r_3\neq i$ are randomly chosen indices. $x_{r_1}^G$ and $x_{r_3}^G$ are selected from $PS^G$, and $x_{r_2}^G$ is chosen from the union of $PS^G$ and the external archive $A$ (i.e., $x_{r_2}^G\in PS^G\cup A^G$). The adaptive best vector $x_{pbest}^G$ is chosen from the top $N_{p}\times p~(p\in [0,1], \text{fixed})$ individuals of generation $G$.
	Inspired from jSO~\cite{brest2017single}, we use the weighted scaling factor $F_w$ defined as:\begin{equation}
		F_w^G=\begin{cases}
		0.7\cdot F^G&\text{if }nfes <= 0.2\cdot nfes_{max},\\
		0.8\cdot F^G&\text{if }nfes <= 0.4\cdot nfes_{max},\\
		1.2\cdot F^G&\text{otherwise}.
		\end{cases}
	\end{equation}
	
	After applying the mutation strategies, the crossover operator is chosen based on CMLwithEN with probability $P_c$, as defined in Section~\ref{subsec: lshade_cnepsin}. A random number $rand(0,1)$ is generated and compared to $P_c$. If $rand(0,1)<P_c$, the CMLwithEN-based crossover is used; otherwise, the binomial crossover is applied.
	\begin{algorithm}
		\caption{Framework of mLSHADE}\label{algo: framework of mLSHADE-cnEpSin}
		\begin{algorithmic}[1]
			\STATE Define the probabilities of all three mutation strategies $P_{MS_i}$, $i=1,2,3$,
			\STATE Use multi-operator mutation strategies as
			\IF{$rand<P_{MS_1}$}
			\STATE Use \textit{DE/current-to-pbest-weight/1 with archive}, 
			\ELSIF{$rand\geq P_{MS_1} ~\&~ rand<P_{MS_2}$}
			\STATE Use \textit{DE/current-to-pbest/1 without archive}, 
			 \STATE Use \textit{DE/current-to-ordpbest-weight/1},
			\ENDIF
			\IF {$rand(0,1)<P_c$}
			\STATE Use CMLwithEN crossover, explained in Section~\ref{subsec: lshade_cnepsin},
			\ELSE 
			\STATE Use binomial crossover, explained in Section~\ref{subsec: de},
			\ENDIF 
			\STATE Update the probability of mutation strategies using Eq.~\eqref{eq: update de probability}.
		\end{algorithmic}
	\end{algorithm}
	\subsubsection{Update Number of Solutions in Each Strategies}\label{subsec: update solutions}
	Initially, the probability of evolving any individual using $MS_1$, $MS_2$, and $MS_3$ is set to $P_{MS_1} = P_{MS_2} =
	P_{MS_3} = 1/3$. For each individual $i$ in $PS_1$, if $rand_i \leq 0.33$, then $x_i^G$ is evolved using $MS_1$. If $0.33\leq rand_i\leq 0.667$, $MS_3$ used. Otherwise, $MS_3$ is applied. To ensure robust performance, the improvement rates in fitness values are used to update the probability of each variant ($P_{MS_i}$)~\cite{elsayed2016testing}. These rates are calculated at the end of each generation as follows:\begin{equation}\label{eq: update de probability}
	\begin{cases}
	P_{MS_i}=max\left(0.1,min\left(0.9,\frac{I_{MS_i}}{\sum_{i=1}^3I_{MS_i}}\right)\right),\\
	I_{MS_i}=\frac{\sum_{i=1}^{PS_1}max(0,f_{v,new_i}-f_{v,old_i})}{\sum_{i=1,MS=i}^{PS_1}f_{v,old_i}},\\
	x_i~\text{evolve by}~ MS_i,
	\end{cases}	,
	\end{equation}where $f_{v,old}$ and $f_{v,new}$ are the old and new fitness values, respectively. Since an operator may perform well at different stages of the evolutionary process and poorly at others, a minimum value of $P_{MS_i}$ is maintained to ensure that each variant has a chance to improve~\cite{elsayed2016testing}.

	\subsubsection{Adaptation of $F$ and $CR$}\label{subsec: parameter adaptation}
	An ensemble of parameter adaptation methods~\cite{awad2016ensemble} adjusts the scaling factor $F$ using two sinusoidal strategies, defined in Eq. \eqref{eq: ensemble lshade-epsin} of Section \ref{subsec: lshade-epsin}.  Initially, both strategies have equal probability $P_j$ and are chosen randomly for the first $L_p$ generations. Subsequently, probabilities are updated using Eq.~\eqref{eq: effective selection}, favoring the strategy with the higher probability. Both strategies are used in the first half of the generations. In the latter half, the scaling factor $F_{i}^G$ is updated using Cauchy distribution:\begin{equation}
	F_i^G=randc(\mu F_{r_i},0.1).
	\end{equation}
	
	The crossover rate $CR_i^G$ is adapted using a normal distribution in the whole evolutionary procedure: \begin{equation}
	CR_i^G=randn(\mu CR_{r_i},0.1).
	\end{equation} If a trial vector $u_i^G$ survives after competing with $x_i^G$, the corresponding $F_i^G$ and $CR_i^G$ values are recorded as successful ($S_F$ and $S_{CR}$). At the end of the generation, the $F$ and $CR$ memories ($\mu F_{r_i}^G$ and $\mu CR_{r_i}^G$) are updated using the Lehmer mean ($mean_{WL}$):\begin{align}
	&\begin{cases}\mu F_{r_i}^{G+1}=\begin{cases}
	mean_{WL}(S_F),&\text{if } S_F\neq \O,\\
	\mu F_{r_i}^G,&\text{otherwise},
	\end{cases}\\\mu CR_{r_i}^{G+1}=\begin{cases}
	mean_{WL}(S_{CR}),&\text{if } S_{CR}\neq \O,\\
	\mu CR_{r_i}^G,&\text{otherwise},
	\end{cases}\end{cases}\\\nonumber&
	mean_{WL}(S)=\frac{\sum_{i=1}^{|S|}\Omega_i\cdot S^2_i}{\sum_{i=1}^{|S|}\Omega_i\cdot S_i},
	\Omega_i=\frac{|f_v(u_i^G)-f_v(x_i^G)|}{\sum_{i=1}^{|S|}|f_v(u_i^G)-f_v(x_i^G)|}.
	\end{align}
\subsubsection{Population Size Reduction}\label{subsec: population reduction}
To maintain diversity during the early stages of the evolutionary process while enhancing exploitation in later stages~\cite{tanabe2014improving}, a linear reduction method is applied to update the size of the population $PS$ at the end of each generation by removing the worst individual. The LPSR method is detailed in Eq. \eqref{eq: linear reduction pop} in Section \ref{subsec: lshade}.
\subsection{Restart Mechanism}\label{subsec: restart mechanism}
As discussed in~\cite{song2023differential}, DE stagnates during evolution, particularly on multimodal optimization problems. This issue often arises due to a lack of population diversity in the later phases of evolution. To address this, we integrate a mechanism to enhance population diversity, which consists of two parts: detecting stagnating individuals and enhancing population diversity. 
\subsubsection{Detection of Stagnating Individuals} 
The evolution process stabilizes after a certain number of generations, and some individuals fail to improve over several generations, leading to stagnation~\cite{song2023differential}. To record the stagnancy, this paper uses counter and diversity metrics. When both conditions are satisfied, we call it stagnation. The counter for each individual is initially defined as zero and updated when the offspring is not better than the parent (Algorithm~\ref{algo: counters}).\begin{algorithm}
	\caption{Record the counters}\label{algo: counters}
	\begin{algorithmic}[1]
	\FOR{$i=1:N_p$}
	\IF{$f_v(u_i^G)>f_v(x_i^G)$}
	\STATE $count(i)=count(i)+1$,
	\ELSE \STATE $count(i)=0$.
	\ENDIF
	\ENDFOR
	\end{algorithmic}
\end{algorithm} 

 The square root of the ratio of the volume associated with the boundaries of the search space $Vol_{bnd}$ and the volume of the population $Vol_{pop}$ in each generation are used to calculate the diversity metrics, as follows:\begin{equation}
	Vol=\sqrt{Vol_{pop}/Vol_{bnd}},
\end{equation}where $Vol_{bnd}=\sqrt{\prod_{j=1}^{D}(x_{j,ub_j}-x_{j,lb})}$ and $Vol_{pop}=\sqrt{\sum_{j=1}^{D}(max(x_{j})-min(x_{j}))/2}$. 

\subsubsection{Replacement of the Stagnating Individuals} Horizontal and vertical crossovers, inspired by genetic algorithms, form a simple competition mechanism to enhance global search capability and reduce blind spots. Horizontal crossover splits the solution space into semi-group hypercubes, performing edge searches to improve global search capability. Individuals of the same dimension are randomly pre-screened and paired for updates, which can be repeated. Vertical crossover allows crossover operations in different sizes, helping to avoid local optima in specific dimensions without affecting others.

The horizontal crossover is defined as:\begin{equation}\label{eq: horizontal}
	Hx_{i_1,D}^G=rd_1\cdot x_{i_1,D}^G+rd_2\cdot x_{i_2,D}^G+rnds\cdot (x_{i_1,D}^G-x_{i_2,D}^G),
\end{equation}where $x_{i_1}^G$ and $x_{i_2}^G$ are two individuals randomly selected from the population. $rd_1$ ($rd_2=1-rd_1$) and $rnds$ are random numbers distributed within $[0,1]$ and $[-1,1]$, respectively. 

The vertical crossover is defined as:\begin{equation}\label{eq: vertical}
	Vx_{i,d_1}^G=rd_1\cdot x_{i,d_1}^G+rd_2\cdot x_{i,d_2}^G,
\end{equation} where $d_1$ and $d_2$ ($d_1,~d_2\in [1,D]$) are different dimensions of a certain stagnant individual. A detailed of the restart mechanism is given in Algorithm~\ref{algo: restart mechanism}.
\begin{algorithm}
	\caption{Restart mechanism}\label{algo: restart mechanism}
	\begin{algorithmic}[1]
		\FOR{$i=1:N_p$}
		\IF{$count(i)>2\cdot D~\&\&~Vol<0.001$}
		\IF{$rand(0,1)>0.5$}
		\STATE Use horizontal crossover, defined in Eq.~\eqref{eq: horizontal},
		\ELSE \STATE Use vertical crossover, defined in Eq.~\eqref{eq: vertical}.
		\ENDIF
		\ENDIF
		\ENDFOR
	\end{algorithmic}
\end{algorithm} 

	\subsection{Local Search}\label{subsec: local search}
	To enhance the exploitation capability of mLSHADE-RL, the SQP method is applied to the best individual found so far during the last 25$\%$ of the evolutionary process. This is done with a probability of $P_{LS}=0.1$ and for up to $nfes_{LS}$ fitness evaluations. Notably, $P_{LS}$ is dynamic: if the local search does not find a better solution, $P_{LS}$ is reduced to a small value, such as 0.01; otherwise, it remains at 0.1.
	\begin{algorithm}
		\caption{Local search}\label{algo: local search}
		\begin{algorithmic}[1]
			\STATE Define the local search probability $P_{LS}$,
			\IF{$rand(0,1)\leq P_{LS}$}
			\STATE Apply the SQP method to find the best possible solution for up to $nfes_{LS}$ fitness evaluations,
			\IF{improvement $\uparrow$}
			\STATE $P_{LS}=0.1$ and update the best solution in $PS$,
			\ELSE \STATE $P_{LS}=0.01$.
			\ENDIF
			\ENDIF
		\end{algorithmic}
	\end{algorithm}

\section{Experimental results and discussions}\label{sec: experimental}
To evaluate the optimization ability of the proposed algorithm, we tested it on the CEC 2024 single-objective bound-constrained problems, which consist of the 30-D problems from the CEC 2017 single-objective optimization test suite~\cite{awad2016problem}. This test suite includes thirty problems with varying characteristics: $cec24_1$ to $cec24_3$ are unimodal, $cec24_4$ to $cec24_{10}$ are multimodal, $cec24_{11}$ to $cec24_{20}$ are hybrid, and the remaining ten are composite problems. More details can be found in~\cite{awad2016problem}.

The proposed mLSHADE-RL was implemented in MATLAB R2024a and ran on a PC with Windows 10 Pro, Intel Core $i7-8700T CPU$, 2.40GHz, and 8 GB RAM. According to the benchmark rules, all algorithms are run 25 times for 10000$*D$ fitness function evaluations for problems with 30$-D$. The error ($|f_v(x_{best}) - f_v(x^*)|$, where $x^*$ is the global optimal solution and $x_{best}$ is the best solution obtained by the proposed algorithm) best, mean, median, worst, and standard deviation (Std) results are recorded. If the distance from the optimal solution is less than or equal to $1E−08$, it is set to zero for that run. We conducted a non-parametric Wilcoxon signed-rank test to compare the statistical algorithms.

  \subsection{Algorithm Parameters} The following are the parameter values of mLSHADE-RL.\begin{enumerate}
\item The initial values of $\mu F_r$, $\mu CR_r$, and $\mu f_r$ are set 0.5. 
\item The memory size is set to 5.  
\item The probabilities $P_c$ and $P_s$	are 0.4 and 0.5~\cite{awad2017ensemble}.
\item The local search probability $P_{LS}$ is 0.01. If it shows an improvement, it is reset by 0.1.
\item The initial population size $N_p^{init}=18*D$ and the minimum population size is 4.
\end{enumerate}
\subsection{Detailed mLSHADE-RL Results}
The detailed results obtained from the proposed mLSHADE-RL are presented in Table~\ref{tab: proposed results}, which includes the best, mean, median, worst, and standard deviation (Std) of error values across 25 runs.

For unimodal problems ($cec24_1$ to $cec24_3$), the algorithm successfully achieves the optimal solution. For multimodal problems ($cec24_4$ to $cec24_{10}$), the algorithm finds the optimal solutions for three out of seven problems. Regarding hybrid problems ($cec24_{11}$ to $cec24_{20}$), the algorithm performs well on all except $cec24_{12}$. For composition problems ($cec24_{21}$ to $cec24_{30}$), which are notably difficult due to a large number of local optima, mLSHADE-RL tends to get stuck. However, the algorithm consistently converges to the same fitness value for several problems.

\begin{table}
\caption{Experimental results of mLSHADE-RL on the CEC 2024 benchmark problems at 30$-D$}\label{tab: proposed results}
\resizebox{1\linewidth}{!}{\begin{tabular}{|lccccc|}\hline
Func	&	Best	&	Worst	&	Median	&	Mean	&	Std	\\\hline
$cec24_1$	&	0.00E+00	&	0.00E+00	&	0.00E+00	&	0.00E+00	&	0.00E+00	\\\hline
$cec24_2$	&	0.00E+00	&	0.00E+00	&	0.00E+00	&	0.00E+00	&	0.00E+00	\\\hline
$cec24_3$	&	0.00E+00	&	0.00E+00	&	0.00E+00	&	0.00E+00	&	0.00E+00	\\\hline
$cec24_4$	&	0.00E+00	&	7.08E+01	&	3.99E+00	&	6.93E+00	&	1.52E+01	\\\hline
$cec24_5$	&	2.98E+00	&	1.59E+01	&	7.96E+00	&	8.08E+00	&	3.14E+00	\\\hline
$cec24_6$	&	0.00E+00	&	4.56E-02	&	0.00E+00	&	2.95E-03	&	1.05E-02	\\\hline
$cec24_7$	&	3.59E+01	&	4.85E+01	&	3.90E+01	&	3.98E+01	&	3.17E+00	\\\hline
$cec24_8$	&	3.72E+00	&	1.59E+01	&	7.96E+00	&	7.95E+00	&	2.66E+00	\\\hline
$cec24_9$	&	0.00E+00	&	0.00E+00	&	0.00E+00	&	0.00E+00	&	0.00E+00	\\\hline
$cec24_{10}$	&	1.02E+03	&	2.02E+03	&	1.48E+03	&	1.47E+03	&	2.92E+02	\\\hline
$cec24_{11}$	&	9.95E-01	&	5.86E+01	&	5.97E+00	&	8.12E+00	&	1.10E+01	\\\hline
$cec24_{12}$	&	4.70E+02	&	2.30E+03	&	1.08E+03	&	1.18E+03	&	4.34E+02	\\\hline
$cec24_{13}$	&	2.21E+00	&	3.64E+01	&	2.09E+01	&	1.97E+01	&	8.52E+00	\\\hline
$cec24_{14}$	&	7.51E+00	&	2.60E+01	&	2.31E+01	&	2.28E+01	&	3.42E+00	\\\hline
$cec24_{15}$	&	1.33E+00	&	7.06E+01	&	1.01E+01	&	1.25E+01	&	1.32E+01	\\\hline
$cec24_{16}$	&	3.26E+00	&	1.32E+02	&	1.42E+01	&	5.57E+01	&	5.83E+01	\\\hline
$cec24_{17}$	&	1.40E+01	&	5.28E+01	&	3.42E+01	&	3.60E+01	&	9.18E+00	\\\hline
$cec24_{18}$	&	2.12E+01	&	5.11E+01	&	2.98E+01	&	3.05E+01	&	7.56E+00	\\\hline
$cec24_{19}$	&	5.11E+00	&	2.75E+01	&	9.33E+00	&	1.09E+01	&	4.88E+00	\\\hline
$cec24_{20}$	&	1.71E+01	&	5.92E+01	&	4.03E+01	&	4.15E+01	&	1.04E+01	\\\hline
$cec24_{21}$	&	2.04E+02	&	2.14E+02	&	2.08E+02	&	2.08E+02	&	2.02E+00	\\\hline
$cec24_{22}$	&	1.00E+02	&	1.00E+02	&	1.00E+02	&	1.00E+02	&	0.00E+00	\\\hline
$cec24_{23}$	&	3.40E+02	&	3.66E+02	&	3.59E+02	&	3.57E+02	&	7.50E+00	\\\hline
$cec24_{24}$	&	4.15E+02	&	4.33E+02	&	4.26E+02	&	4.25E+02	&	4.44E+00	\\\hline
$cec24_{25}$	&	3.79E+02	&	3.87E+02	&	3.80E+02	&	3.81E+02	&	2.67E+00	\\\hline
$cec24_{26}$	&	8.54E+02	&	1.13E+03	&	9.89E+02	&	9.91E+02	&	7.57E+01	\\\hline
$cec24_{27}$	&	4.88E+02	&	5.22E+02	&	5.02E+02	&	5.04E+02	&	9.06E+00	\\\hline
$cec24_{28}$	&	3.00E+02	&	3.00E+02	&	3.00E+02	&	3.00E+02	&	4.33E-13	\\\hline
$cec24_{29}$	&	3.65E+02	&	4.53E+02	&	4.37E+02	&	4.27E+02	&	2.06E+01	\\\hline
$cec24_{30}$	&	1.70E+03	&	2.36E+03	&	1.93E+03	&	1.95E+03	&	1.69E+02	\\\hline
\end{tabular}}
\end{table}
\subsection{Comparison with Other State-of-the-art Algorithms}
The performance of mLSHADE-RL is compared with several state-of-the-art algorithms, including the effective butterfly algorithm with CMA-ES (EBOwithCMAR)\cite{kumar2017improving} (first winner of the CEC 2017 competition), LSHADE-cnEpSin\cite{awad2017ensemble} (third place in CEC 2017), LSHADE with semi-parameter adaptation with CMA-ES (LSHADE-SPACMA)\cite{mohamed2017lshade} (fourth place in CEC 2017), hybrid sampling evolution strategy (HS-ES)\cite{zhang2018hybrid} (first winner of the CEC 2018 competition), Enhanced LSHADE-SPACMA (ELSHADE-SPACMA) (third place in CEC 2018), improved multi-operator DE (IMODE)\cite{sallam2020improved}, and the MadDE algorithm that improves DE through Bayesian hyper-parameter optimization\cite{biswas2021improving}. For the comparative algorithms, parameter values were taken from the relevant literature\footnote{The codes of these algorithms are downloaded from \url{https://github.com/P-N-Suganthan}}.

Table~\ref{tab: statistical results} presents the Wilcoxon sign-rank test results of the proposed algorithm with existing algorithms.
Compared with CEC 2017 winners, the proposed mLSHADE-RL algorithm is superior to
EBOwithCMAR, LSHADE-cnEpSin, and LSHADE-SPACMA
in 7, 2, and 5 test problems, respectively, obtained similar results
to them for 15, 15, and 14 test problems and inferior for 8, 13, and 11 test problems, respectively. While with CEC 2018 winners, mLSHADE-RL provides better statistical results than HS-ES and ELSHADE-SPACMA in 9 and 10 issues, similar results in 10 and 13 issues, and inferior results in 11 and 7 issues, respectively. Compared with other algorithms, such as IMODE and MadDE, the proposed algorithm provides better statistical results in 28 problems, similar results in 1 and 0, and inferior results in 1 and 2 issues, respectively.
\begin{table}
	\caption{A comparison between the proposed mLSHADE and other state-of-the-art algorithms based on mean results}\label{tab: statistical results}
	\begin{tabular}{|c|ccc|}\hline
		Algorithms	&	Better	&	Similar	&	Worse	\\\hline
		mLSHADE-RL $vs$ LSHADE-cnEpSin	&	7	&	15	&	8	\\\hline
		mLSHADE-RL $vs$ EBOwithCMAR	&	2	&	15	&	13	\\\hline
		mLSHADE-RL $vs$ LSHADE-SPACMA	&	5	&	14	&	11	\\\hline
		mLSHADE-RL $vs$ HS-ES	&	9	&	10	&	11	\\\hline
		mLSHADE-RL $vs$ ELSHADE-SPACMA	&	10	&	13	&	7	\\\hline
		mLSHADE-RL $vs$ IMODE	&	28	&	1	&	1	\\\hline
		mLSHADE-RL $vs$ MadDE	&	28	&	0	&	2	\\\hline
		
	\end{tabular}
\end{table}
\section{Conclusion}\label{sec: conclusion}
This paper proposes mLSHADE-RL, an enhanced version of LSHADE-cnEpSin, which integrates multiple EAs and operators. The population is updated using a multi-operator LSHADE-cnEpSin (mLSHADE). A stagnation tracking procedure is also integrated into mLSHADE, and a restart mechanism is used to overcome the stagnancy. Additionally, a local search method is applied later in the evolutionary process to enhance mLSHADE-RL's exploitation capability. mLSHADE-RL was benchmarked on the CEC 2024 special session and competitions' problem suite for single-objective bound constrained real-parameter optimization. Compared with other state-of-the-art algorithms, mLSHADE-RL demonstrated superior performance and the ability to obtain high-quality solutions. Future work will include conducting more sensitivity analysis on algorithm parameters to enhance performance further.

	\bibliographystyle{IEEEtran}
	\bibliography{references_ulshade}

\begin{thebibliography}{10}
\providecommand{\url}[1]{#1}
\csname url@samestyle\endcsname
\providecommand{\newblock}{\relax}
\providecommand{\bibinfo}[2]{#2}
\providecommand{\BIBentrySTDinterwordspacing}{\spaceskip=0pt\relax}
\providecommand{\BIBentryALTinterwordstretchfactor}{4}
\providecommand{\BIBentryALTinterwordspacing}{\spaceskip=\fontdimen2\font plus
\BIBentryALTinterwordstretchfactor\fontdimen3\font minus \fontdimen4\font\relax}
\providecommand{\BIBforeignlanguage}[2]{{%
\expandafter\ifx\csname l@#1\endcsname\relax
\typeout{** WARNING: IEEEtran.bst: No hyphenation pattern has been}%
\typeout{** loaded for the language `#1'. Using the pattern for}%
\typeout{** the default language instead.}%
\else
\language=\csname l@#1\endcsname
\fi
#2}}
\providecommand{\BIBdecl}{\relax}
\BIBdecl

\bibitem{storn1997differential}
R.~Storn and K.~Price, ``Differential evolution--a simple and efficient heuristic for global optimization over continuous spaces,'' \emph{Journal of global optimization}, vol.~11, pp. 341--359, 1997.

\bibitem{das2016recent}
S.~Das, S.~S. Mullick, and P.~N. Suganthan, ``Recent advances in differential evolution--an updated survey,'' \emph{Swarm and evolutionary computation}, vol.~27, pp. 1--30, 2016.

\bibitem{qin2008differential}
A.~K. Qin, V.~L. Huang, and P.~N. Suganthan, ``Differential evolution algorithm with strategy adaptation for global numerical optimization,'' \emph{IEEE transactions on Evolutionary Computation}, vol.~13, no.~2, pp. 398--417, 2008.

\bibitem{zamuda2012population}
A.~Zamuda and J.~Brest, ``Population reduction differential evolution with multiple mutation strategies in real world industry challenges,'' in \emph{International Symposium on Evolutionary Computation}.\hskip 1em plus 0.5em minus 0.4em\relax Springer, 2012, pp. 154--161.

\bibitem{brest2013real}
J.~Brest, B.~Bo{\v{s}}kovi{\'c}, A.~Zamuda, I.~Fister, and E.~Mezura-Montes, ``Real parameter single objective optimization using self-adaptive differential evolution algorithm with more strategies,'' in \emph{2013 IEEE Congress on Evolutionary Computation}.\hskip 1em plus 0.5em minus 0.4em\relax IEEE, 2013, pp. 377--383.

\bibitem{pant2020differential}
M.~Pant, H.~Zaheer, L.~Garcia-Hernandez, A.~Abraham \emph{et~al.}, ``Differential evolution: A review of more than two decades of research,'' \emph{Engineering Applications of Artificial Intelligence}, vol.~90, p. 103479, 2020.

\bibitem{awad2016ensemble}
N.~H. Awad, M.~Z. Ali, P.~N. Suganthan, and R.~G. Reynolds, ``An ensemble sinusoidal parameter adaptation incorporated with l-shade for solving cec2014 benchmark problems,'' in \emph{2016 IEEE congress on evolutionary computation (CEC)}.\hskip 1em plus 0.5em minus 0.4em\relax IEEE, 2016, pp. 2958--2965.

\bibitem{trivedi2017unified}
A.~Trivedi, K.~Sanyal, P.~Verma, and D.~Srinivasan, ``A unified differential evolution algorithm for constrained optimization problems,'' in \emph{2017 IEEE congress on evolutionary computation (CEC)}.\hskip 1em plus 0.5em minus 0.4em\relax IEEE, 2017, pp. 1231--1238.

\bibitem{tanabe2013success}
R.~Tanabe and A.~Fukunaga, ``Success-history based parameter adaptation for differential evolution,'' in \emph{2013 IEEE congress on evolutionary computation}.\hskip 1em plus 0.5em minus 0.4em\relax IEEE, 2013, pp. 71--78.

\bibitem{zhang2009jade}
J.~Zhang and A.~C. Sanderson, ``Jade: adaptive differential evolution with optional external archive,'' \emph{IEEE Transactions on evolutionary computation}, vol.~13, no.~5, pp. 945--958, 2009.

\bibitem{mohamed2019novel}
A.~W. Mohamed, A.~A. Hadi, and K.~M. Jambi, ``Novel mutation strategy for enhancing shade and lshade algorithms for global numerical optimization,'' \emph{Swarm and Evolutionary Computation}, vol.~50, p. 100455, 2019.

\bibitem{tanabe2014improving}
R.~Tanabe and A.~S. Fukunaga, ``Improving the search performance of shade using linear population size reduction,'' in \emph{2014 IEEE congress on evolutionary computation (CEC)}.\hskip 1em plus 0.5em minus 0.4em\relax IEEE, 2014, pp. 1658--1665.

\bibitem{awad2017ensemble}
N.~H. Awad, M.~Z. Ali, and P.~N. Suganthan, ``Ensemble sinusoidal differential covariance matrix adaptation with euclidean neighborhood for solving cec2017 benchmark problems,'' in \emph{2017 IEEE congress on evolutionary computation (CEC)}.\hskip 1em plus 0.5em minus 0.4em\relax IEEE, 2017, pp. 372--379.

\bibitem{elsayed2011multi}
S.~M. Elsayed, R.~A. Sarker, and D.~L. Essam, ``Multi-operator based evolutionary algorithms for solving constrained optimization problems,'' \emph{Computers \& operations research}, vol.~38, no.~12, pp. 1877--1896, 2011.

\bibitem{sallam2017landscape}
K.~M. Sallam, S.~M. Elsayed, R.~A. Sarker, and D.~L. Essam, ``Landscape-based adaptive operator selection mechanism for differential evolution,'' \emph{Information Sciences}, vol. 418, pp. 383--404, 2017.

\bibitem{tasgetiren2010ensemble}
M.~F. Tasgetiren, P.~N. Suganthan, and Q.-K. Pan, ``An ensemble of discrete differential evolution algorithms for solving the generalized traveling salesman problem,'' \emph{Applied Mathematics and Computation}, vol. 215, no.~9, pp. 3356--3368, 2010.

\bibitem{wu2018ensemble}
G.~Wu, X.~Shen, H.~Li, H.~Chen, A.~Lin, and P.~N. Suganthan, ``Ensemble of differential evolution variants,'' \emph{Information Sciences}, vol. 423, pp. 172--186, 2018.

\bibitem{chen2019united}
J.~Chen, J.~Chen, and H.~Min, ``A united framework with multi-operator evolutionary algorithms and interior point method for efficient single objective optimisation problem solving,'' \emph{International Journal of High Performance Computing and Networking}, vol.~13, no.~3, pp. 340--353, 2019.

\bibitem{elsayed2014testing}
S.~M. Elsayed, R.~A. Sarker, D.~L. Essam, and N.~M. Hamza, ``Testing united multi-operator evolutionary algorithms on the cec2014 real-parameter numerical optimization,'' in \emph{2014 IEEE congress on evolutionary computation (CEC)}.\hskip 1em plus 0.5em minus 0.4em\relax IEEE, 2014, pp. 1650--1657.

\bibitem{elsayed2016testing}
S.~Elsayed, N.~Hamza, and R.~Sarker, ``Testing united multi-operator evolutionary algorithms-ii on single objective optimization problems,'' in \emph{2016 IEEE congress on evolutionary computation (CEC)}.\hskip 1em plus 0.5em minus 0.4em\relax IEEE, 2016, pp. 2966--2973.

\bibitem{kumar2017improving}
A.~Kumar, R.~K. Misra, and D.~Singh, ``Improving the local search capability of effective butterfly optimizer using covariance matrix adapted retreat phase,'' in \emph{2017 IEEE congress on evolutionary computation (CEC)}.\hskip 1em plus 0.5em minus 0.4em\relax IEEE, 2017, pp. 1835--1842.

\bibitem{brest2017single}
J.~Brest, M.~S. Mau{\v{c}}ec, and B.~Bo{\v{s}}kovi{\'c}, ``Single objective real-parameter optimization: Algorithm jso,'' in \emph{2017 IEEE congress on evolutionary computation (CEC)}.\hskip 1em plus 0.5em minus 0.4em\relax IEEE, 2017, pp. 1311--1318.

\bibitem{song2023differential}
Z.~Song and Z.~Meng, ``Differential evolution with wavelet basis function based parameter control and dimensional interchange for diversity enhancement,'' \emph{Applied Soft Computing}, vol. 144, p. 110492, 2023.

\bibitem{awad2016problem}
N.~Awad, M.~Ali, J.~Liang, B.~Qu, and P.~Suganthan, ``Problem definitions and evaluation criteria for the cec 2017 special session and competition on single objective bound constrained real-parameter numerical optimization,'' in \emph{Technical report}.\hskip 1em plus 0.5em minus 0.4em\relax Nanyang Technological University Singapore Singapore, 2016, pp. 1--34.

\bibitem{mohamed2017lshade}
A.~W. Mohamed, A.~A. Hadi, A.~M. Fattouh, and K.~M. Jambi, ``Lshade with semi-parameter adaptation hybrid with cma-es for solving cec 2017 benchmark problems,'' in \emph{2017 IEEE Congress on evolutionary computation (CEC)}.\hskip 1em plus 0.5em minus 0.4em\relax IEEE, 2017, pp. 145--152.

\bibitem{zhang2018hybrid}
G.~Zhang and Y.~Shi, ``Hybrid sampling evolution strategy for solving single objective bound constrained problems,'' in \emph{2018 IEEE Congress on Evolutionary Computation (CEC)}.\hskip 1em plus 0.5em minus 0.4em\relax IEEE, 2018, pp. 1--7.

\bibitem{sallam2020improved}
K.~M. Sallam, S.~M. Elsayed, R.~K. Chakrabortty, and M.~J. Ryan, ``Improved multi-operator differential evolution algorithm for solving unconstrained problems,'' in \emph{2020 IEEE congress on evolutionary computation (CEC)}.\hskip 1em plus 0.5em minus 0.4em\relax IEEE, 2020, pp. 1--8.

\bibitem{biswas2021improving}
S.~Biswas, D.~Saha, S.~De, A.~D. Cobb, S.~Das, and B.~A. Jalaian, ``Improving differential evolution through bayesian hyperparameter optimization,'' in \emph{2021 IEEE congress on evolutionary computation (CEC)}.\hskip 1em plus 0.5em minus 0.4em\relax IEEE, 2021, pp. 832--840.

\end{thebibliography}
\end{document}